\documentclass[11pt,a4paper]{article}
\usepackage[hyperref]{emnlp-ijcnlp-2019}
\usepackage{times}
\usepackage{latexsym}

\usepackage{url}

\usepackage{amsfonts}
\usepackage{amsmath}
\usepackage{booktabs}
\usepackage{multirow}
\usepackage{graphicx}
\usepackage{enumitem}
\usepackage{textcomp}

\aclfinalcopy 

\usepackage{multirow}
\usepackage{graphicx}
\usepackage{booktabs}
\usepackage{enumitem}
\usepackage{amsmath}
\usepackage{amsfonts}

\newcommand*{\affaddr}[1]{#1}
\newcommand*{\affmark}[1][*]{\textsuperscript{#1}}
\newcommand*{\email}[1]{#1}

\title{A Split-and-Recombine Approach for Follow-up Query Analysis}

\author{Qian Liu\affmark[\textdagger]{\thanks{~~Work done during an internship at Microsoft Research.}}~~, Bei Chen\affmark[\S], Haoyan Liu\affmark[$\lozenge$]$^*$, Jian-Guang Lou\affmark[\S], Lei Fang\affmark[\S],\\
\textbf{Bin Zhou\affmark[\textdagger], Dongmei Zhang\affmark[\S]}\\
\affaddr{\affmark[\textdagger]State Key Laboratory of Virtual Reality Technology and Systems,\\ School of Computer Science and Engineering, Beihang University, China}\\
\affaddr{\affmark[$\lozenge$]State Key Lab of Software Development Environment, Beihang University, China}\\
\affaddr{\affmark[\S]Microsoft Research, Beijing, China}\\
\affmark[\textdagger]\affmark[$\lozenge$]\email{\{qian.liu, haoyan.liu, zhoubin\}@buaa.edu.cn;\\ \affmark[\S]\{beichen, jlou, leifa, dongmeiz\}@microsoft.com}\\
}

\date{}

\begin{document}
\maketitle
\begin{abstract}

Context-dependent semantic parsing has proven to be an important yet challenging task. To leverage the advances in context-independent semantic parsing, we propose to perform follow-up query analysis, aiming to restate context-dependent natural language queries with contextual information. To accomplish the task, we propose \textsc{StAR}, a novel approach with a well-designed two-phase process. It is parser-independent and able to handle multifarious follow-up scenarios in different domains. Experiments on the FollowUp dataset show that \textsc{StAR} outperforms the state-of-the-art baseline by a large margin of nearly $8\%$. The superiority on parsing results verifies the feasibility of follow-up query analysis. We also explore the extensibility of \textsc{StAR} on the SQA dataset, which is very promising.

\end{abstract}

\section{Introduction}

Recently, Natural Language Interfaces to Data- bases (NLIDB) has received considerable attention, as they allow users to query databases by directly using natural language. Current studies mainly focus on context-independent semantic parsing, which translates a single natural language sentence into its corresponding executable form (e.g. Structured Query Language) and retrieves the answer from databases regardless of its context. However, context does matter in real world applications. Users tend to issue queries in a coherent way when communicating with NLIDB. For example, after the query ``How much money has Smith earned?'' (\textbf{Precedent Query}), users may pose another query by simply asking ``How about Bill Collins?'' (\textbf{Follow-up Query}) instead of the complete ``How much money has Bill Collins earned?'' (\textbf{Restated Query}). Therefore, contextual information is essential for more accurate and robust semantic parsing, namely context-dependent semantic parsing.

Compared with context-independent semantic parsing, context-dependent semantic parsing has received less attention. Several attempts include a statistical model with parser trees \cite{miller1996fully}, a linear model with context-dependent logical forms \cite{zettlemoyer2009learning} and a sequence-to-sequence model \cite{suhr2018learning}. However, all these methods cannot apply to different domains, since the ATIS dataset \cite{dahl1994expanding} they rely on is domain-specific. A search-based neural method DynSP$^*$ arises along with the SequentialQA~(SQA) dataset~\cite{iyyer2017search}, which takes the first step towards cross-domain context-dependent semantic parsing. Nevertheless, DynSP$^*$ focuses on dealing with relatively simple scenarios. All the aforementioned methods design context-dependent semantic parser from scratch. Instead, inspired by~\citet{liu2019fanda}, we propose to directly leverage the technical advances in context-independent semantic parsing. We define \emph{follow-up query analysis} as restating the follow-up queries using contextual information in natural language, then the restated queries can be translated to the corresponding executable forms by existing context-independent parsers. In this way, we boost the performance of context-dependent semantic parsing.

In this paper, we focus on follow-up query analysis and present a novel approach. The main idea is to decompose the task into two phases by introducing a learnable intermediate structure \emph{span}: two queries first get split into several spans, and then undergo the recombination process. As no intermediate annotation is involved, we design rewards to jointly train the two phases by applying reinforcement learning (RL) \cite{Sutton1998ReinforcementIntroduction}. Our major contributions are as follows:

\begin{itemize}[leftmargin=*]\setlength\itemsep{-0.3em}
    \item We propose a novel approach, named \textbf{S}pli\textbf{T}-\textbf{A}nd-\textbf{R}ecombine (\textsc{StAR})\footnote{Code is available at {\small \url{http://github.com/microsoft/EMNLP2019-Split-And-Recombine}}.}, to restate follow-up queries via two phases. It is parser-independent and can be seamlessly integrated with existing context-independent semantic parsers.
    \item We conduct experiments on the FollowUp dataset \cite{liu2019fanda}, which covers multifarious cross-domain follow-up scenarios. The results demonstrate that our approach significantly outperforms the state-of-the-art baseline.
    \item We redesign the recombination process and extend \textsc{StAR} to the SQA dataset, where the annotations are answers. Experiments show promising results, that demonstrates the extensibility of our approach.
\end{itemize}

\section{Methodology}

In this section, we first give an overview of our proposed method with the idea of two-phase process, then introduce the two phases in turn.

\subsection{Overview of Split-And-Recombine}\label{sec-overview}

Let $\mathbf{x}=(x_1,\dots,x_n)$, $\mathbf{y}=(y_1,\dots,y_m)$ and $\mathbf{z}=(z_1,\dots,z_l)$ denote the precedent query, follow-up query and restated query respectively, each of which is a natural language sentence. Our goal is to interpret the follow-up query $\mathbf{y}$ with its precedent query $\mathbf{x}$ as context, and generate the corresponding restated query $\mathbf{z}$. The restated query has the same meaning with the follow-up query, but it is complete and unambiguous to facilitate better downstream parsing. Formally, given the pair $(\mathbf{x}, \mathbf{y})$, we aim to learn a model $P_{\rm{model}}(\mathbf{z}|\mathbf{x}, \mathbf{y})$ and maximize the objective:
\begin{equation}
    \mathcal{L}=\mathbb{E}_{(\mathbf{x}, \mathbf{y}, \mathbf{z}){\sim}\mathcal{D}}[\log P_{\rm{model}}(\mathbf{z}|\mathbf{x}, \mathbf{y})],
\end{equation}
where $\mathcal{D}$ represents the set of training data. As to $P_{\rm{model}}(\mathbf{z}|\mathbf{x}, \mathbf{y})$, since $\mathbf{z}$ always overlaps a great with $\mathbf{x}$ and $\mathbf{y}$, it is intuitively more straightforward to find a way to merge $\mathbf{x}$ and $\mathbf{y}$. To this end, we design a two-phase process and present a novel approach \textsc{StAR} to perform follow-up query analysis with reinforcement learning.

\begin{figure}[t]
    \centering
    \includegraphics[width=.48\textwidth]{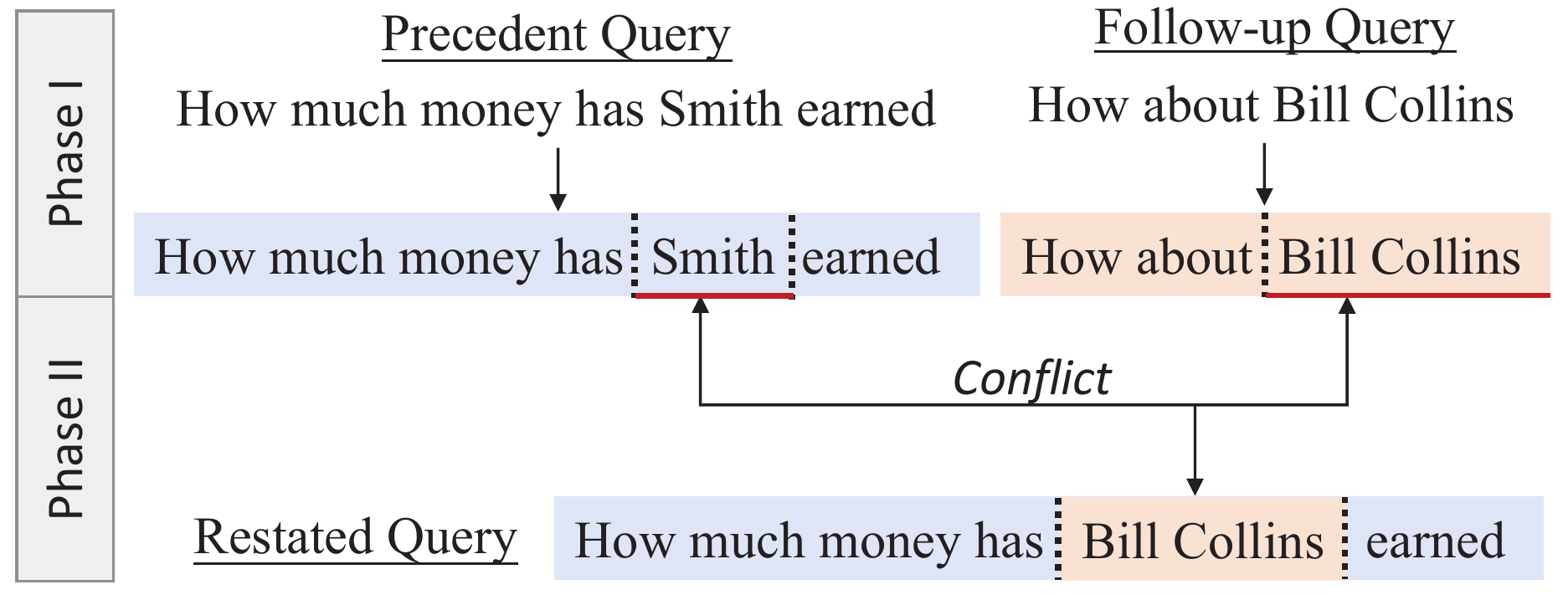}
    \caption{The two-phase process of an example from the FollowUp dataset (More real cases of diverse follow-up scenarios can be found in Table~\ref{table:case_study}).}
    \label{fig:two_phase}
\end{figure}

\begin{figure*}
    \centering
    \includegraphics[width=0.9\linewidth]{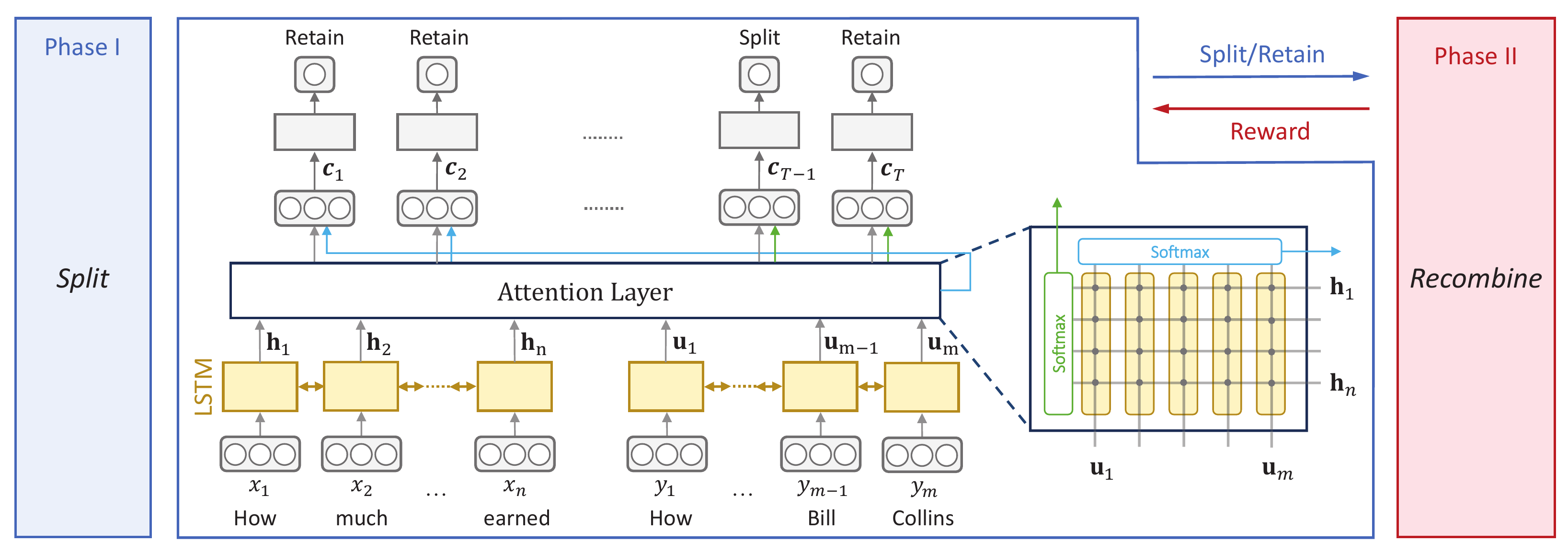}
    \caption{The overview of \textsc{StAR} with two phases.}
    \label{fig:model_framework}
\end{figure*}

A concrete example of the two-phase process is shown in Figure \ref{fig:two_phase}. Phase I is to \textbf{Split} input queries into several spans. For example, the precedent query is split into $3$ spans: ``How much money has'', ``Smith'' and ``earned''. Let $q$ denote a kind of way to split $(\mathbf{x}, \mathbf{y})$, then Phase I can be formulated as $P_{\rm{split}}(q|\mathbf{x}, \mathbf{y})$. Phase II is to \textbf{Recombine} the spans by finding out the most probable \emph{conflicting way}, and generating the final output by restatement, denoted as $P_{\rm{rec}}(\mathbf{z}|q)$. Two spans being conflicting means they are semantically similar. For example, ``Smith'' conflicts with ``Bill Collins''. A conflicting way contains all conflicts between the precedent and follow-up spans. Backed by the two-phase idea of splitting and recombination, the overall likelihood of generating $\mathbf{z}$ given $\mathbf{x},\mathbf{y}$ is:
\begin{equation}
    \!P_{\rm{model}}(\mathbf{z}|\mathbf{x}, \mathbf{y}) = {\sum\limits_{q\in\mathcal{Q}}} P_{\rm{split}}(q|\mathbf{x}, \mathbf{y})P_{\rm{rec}}(\mathbf{z}|q),\!\!
\end{equation}
where $\mathcal{Q}$ represents the set of all possible ways to split $(\mathbf{x}, \mathbf{y})$. Due to the lack of annotations for splitting and recombination, it is hard to directly perform supervised learning. Inspired by \citet{liang2016neural}, we employ RL to optimize $P_{\rm{model}}$. Denoting the predicted restated query by $\tilde{\mathbf{z}}$, simplifying $\mathbb{E}_{(\mathbf{x}, \mathbf{y}, \mathbf{z}){\sim}\mathcal{D}}$ as $\mathbb{E}$, the goal of the RL training is to maximize following objective:
\begin{equation}\label{eq:lrl}
    \!\mathcal{L}_{\rm{rl}}\!=\!\mathbb{E}[{\sum\limits_{\tilde{\mathbf{z}}{\in}\mathcal{Z}}}{\sum\limits_{q\in\mathcal{Q}}}\!P_{\rm{split}}\!(q|\mathbf{x}, \mathbf{y})P_{\rm{rec}}\!(\tilde{\mathbf{z}}|q)r(\mathbf{z},\tilde{\mathbf{z}})],\!\!\!\!\!
\end{equation}
where $\mathcal{Z}$ is the space of all restated query candidates and $r$ represents the reward defined by comparing $\tilde{\mathbf{z}}$ and the annotation $\mathbf{z}$. However, the overall candidate space $\mathcal{Q}{\times}\mathcal{Z}$ is vast, making it impossible to exactly maximize $\mathcal{L}_{\rm{rl}}$. The most straightforward usage of the REINFORCE algorithm \cite{williams1992simple}, sampling both $q$ and $\tilde{\mathbf{z}}$, also poses challenges for learning. To alleviate the problem, we propose to sample $q$ and enumerate all candidate $\tilde{\mathbf{z}}$ after $q$ is determined. It could shrink the sampling space with an acceptable computational cost, which will be discussed in Section~\ref{subsec:variant}. Thus the problem turns to design a reward function $R(q,\mathbf{z})$ to evaluate $q$ and guide the learning. To achieve it, we reformulate Equation~\ref{eq:lrl} as:
\begin{equation}
    \!\mathcal{L}_{\rm{rl}}\!=\!\mathbb{E}[{\sum\limits_{q\in\mathcal{Q}}}\!P_{\rm{split}}(q|\mathbf{x}, \mathbf{y})\!{\sum\limits_{\tilde{\mathbf{z}}{\in}\mathcal{Z}}}\!P_{\rm{rec}}(\tilde{\mathbf{z}}|q)r(\mathbf{z},\tilde{\mathbf{z}})], \!\!\!\!\!
\end{equation}
and set the $R(q,\mathbf{z})$ as:
\begin{equation}\label{eq:rs}
{\sum\limits_{\tilde{\mathbf{z}}{\in}\mathcal{Z}}}P_{\rm{rec}}(\tilde{\mathbf{z}}|q)r(\mathbf{z},\tilde{\mathbf{z}}).
\end{equation}
The overview of \textsc{StAR} is summarized in Figure \ref{fig:model_framework}. Given $\mathbf{x},\mathbf{y}$, during training of Phase I (in blue), we fix $P_{\rm{rec}}$ to provide the reward $R(q,\mathbf{z})$, then $P_{\rm{split}}$ can be learnt by the REINFORCE algorithm. During training of Phase II (in red), we fix $P_{\rm{split}}$ and utilize it to generate $q$, $P_{\rm{rec}}$ is trained to maximize Equation~\ref{eq:rs}. In this way, $P_{\rm{split}}$ and $P_{\rm{rec}}$ can be jointly trained. The details are introduced below.

\subsection{Phase I: Split}

\begin{figure*}
    \centering
    \includegraphics[width=.8\textwidth]{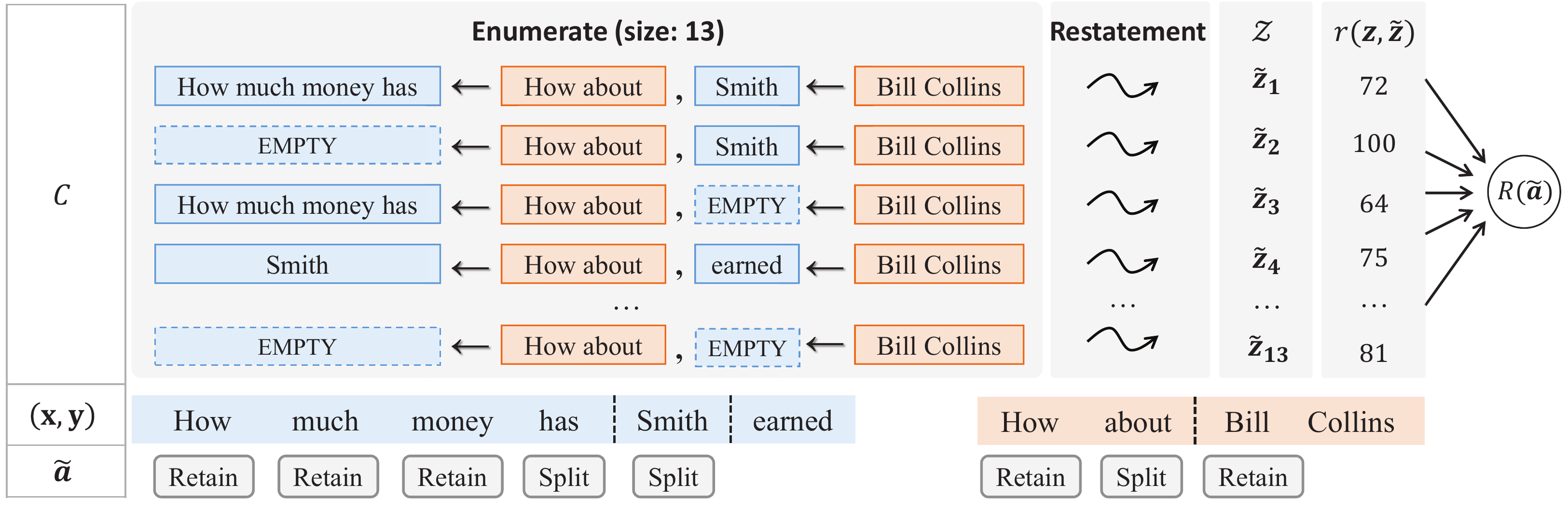}
    \caption{Illustration of reward computation in Phase II.}
    \label{fig:span_restate}
\end{figure*}

As mentioned above, fixed $P_{\rm{rec}}$, Phase I updates $P_{\rm{split}}$, the Split Neural Network (\textbf{SplitNet}). Taking the precedent query and follow-up query as input, as shown in Figure \ref{fig:model_framework}, splitting spans can be viewed as a sequence labeling problem over input. For each word, SplitNet outputs a label \emph{Split} or \emph{Retain}, indicating whether a split operation will be performed after the corresponding word. A label sequence uniquely identifies a way of splitting $(\mathbf{x}, \mathbf{y})$, mentioned as $q$ in Section \ref{sec-overview}. Figure \ref{fig:span_restate} gives an example on the bottom. In the precedent query, two split operations are performed after ``has'' and ``Smith'' , since their labels are \emph{Split}.

\subsubsection{Split Neural Network}\label{sec:state}

Intuitively, only after obtaining information from both the precedent query and follow-up query can SplitNet get to know the reasonable way to split spans. Inspired by BiDAF~\cite{seo2016bidirectional}, we apply a bidirectional attention mechanism to capture the interrelations between the two queries.

\paragraph{Embedding Layer} We consider embedding in three levels: character, word and sentence, respectively denoted as $\phi_c$, $\phi_w$ and $\phi_s$. Character-level embedding maps each word to a vector in a high-dimensional space using Convolutional Neural Networks \cite{kim2014convolutional}. Word-level embedding is initialized using GloVe \cite{pennington2014glove}, and then it is updated along with other parameters. Sentence-level embedding is a one-hot vector designed to distinguish between precedent and follow-up queries. Then, the overall embedding function is $\phi=[\phi_c; \phi_w; \phi_s]$.

\paragraph{Context Layer} On top of the embedding layer, Bidirectional Long Short-Term Memory Network (BiLSTM) \cite{hochreiter1997long,schuster1997bidirectional} is applied to capture contextual information within one query. For word $x_i(i\!=\!1,\dots,n)$ in the precedent query $\mathbf{x}$, the hidden state $\mathbf{h}_i=[\mathop{\overrightarrow{\mathbf{h}}_i};\overleftarrow{\mathbf{h}}_i]$ is computed, where the forward hidden state is:
\begin{equation}
\overrightarrow{\mathbf{h}}_i = {\overrightarrow{\mathbf{LSTM}}} \big( \phi(x_i);\overrightarrow{\mathbf{h}}_{i-1} \big).
\end{equation}
Similarly, a hidden state $\mathbf{u}_j$ is computed for word $y_j(j\!=\!1,\dots,m)$. The BiLSTMs for $\mathbf{x}$ and $\mathbf{y}$ share the same parameters.

\paragraph{Attention Layer} The interrelations between the precedent and follow-up queries are captured via attention layer. Let $\mathbf{H}\!=\![\mathbf{h}_1,\mathbf{h}_2,\dots,\mathbf{h}_n]$ and $\mathbf{U}\!=\![\mathbf{u}_1,\mathbf{u}_2,\dots,\mathbf{u}_m]$ denote the hidden states of two queries respectively, the similarity matrix is:
\begin{equation}
    \mathbf{A} = \cos(\mathbf{H}^\top \mathbf{U}),
\end{equation}
where $\mathbf{A}\!\in\!\mathbb{R}^{n \times m}$ and the entry $A_{i,j}$ represents the similarity between words $x_i$ and $y_j$. Then the softmax function is used to obtain the precedent-to-follow \textbf{(P2F)} attention and the follow-to-precedent \textbf{(F2P)} attention. P2F attention represents $y_j$ using the similarities between $y_j$ and every word in $\mathbf{x}$. Specifically, let $\mathbf{f}_j={\rm softmax}(\mathbf{A}_{:,j})$, where $\mathbf{f}_j\in\mathbb{R}^n$ denotes the attention weights on $\mathbf{x}$ according to $y_j$. Then $y_j$ can be represented by a precedent-aware vector $\mathbf{\Tilde{u}}_{j}=\sum_{k=1}^n\mathbf{f}_{j}[k]{\cdot}\mathbf{h}_{k}$. Similarly, F2P attention computes the attention weights on $\mathbf{y}$ according to $x_i$, and represents $x_i$ as $\mathbf{\Tilde{h}}_{i}$.

\paragraph{Output Layer}

Combining the outputs of the context layer and the attention layer, we design the final hidden state as follows:
\begin{align}
    \mathbf{c}^{x_i} &= [\mathbf{h}_i;~\mathbf{h}_i{\circ}\mathbf{\Tilde{h}}_i;~\mathbf{h}_{i+1}{\circ}\mathbf{\Tilde{h}}_{i+1}], \\
    \mathbf{c}^{y_j} &= [\mathbf{u}_j;~\mathbf{u}_j{\circ}\mathbf{\Tilde{u}}_j;~\mathbf{u}_{j+1}{\circ}\mathbf{\Tilde{u}}_{j+1}], 
\end{align}
where $\!i\!\!\in\!\!\!\{\!1,\!\dots,\!n\!\!-\!\!\!1\!\}$,$j\!\!\in\!\!\!\{\!1,\!\dots,\!m\!\!-\!\!\!1\!\}\!$ and $\!\circ$\! denotes element-wise multiplication \cite{Lee_2017}. Let $\!\mathbf{c}\!=\!(\mathbf{c}_t)_{t=1}^T\!=\!(\mathbf{c}^{x_1},...,\mathbf{c}^{x_{n-1}},\mathbf{c}^{y_1},...,\mathbf{c}^{y_{m-1}})$ denote the final hidden state sequence. At each position $t$, the probability of \textit{Split} is $\sigma(\mathbf{W}*\mathbf{c}_{t} + b)$, where $\sigma$ denotes the \emph{sigmoid} function and $\{\mathbf{W},b\}$ denotes the parameters.

\subsubsection{Training}

It is difficult to train RL model from scratch. Therefore, we propose to initialize SplitNet via pre-training, and then use reward to optimize it.

\paragraph{Pre-training} We obtain the pre-training annotation $\mathbf{a}$ by finding the common substrings between $(\mathbf{x},\mathbf{y})$ and $\mathbf{z}$. One $\mathbf{a}$ is a label sequence, each of which is \emph{Split} or \emph{Retain}. Given the pre-training data set $\mathcal{D}_{\rm{pre}}$ whose training instance is as $(\mathbf{x}, \mathbf{y}, \mathbf{a})$, the objective function of pre-training is:
\begin{equation}
    \mathcal{L}_{\rm{pre}}(\theta)=\mathbb{E}_{(\mathbf{x}, \mathbf{y}, \mathbf{a}){\sim}\mathcal{D}_{\rm{pre}}}[\log p_{\theta}(\mathbf{a}|\mathbf{x},\mathbf{y})],
\end{equation}
where $\theta$ is the parameter of SplitNet.

\paragraph{Policy Gradient} After pre-training, we treat the label sequence as a variable $\tilde{\mathbf{a}}$. The reward $R(\tilde{\mathbf{a}},\mathbf{z})$ (details in Section \ref{SRS}) is used to optimize the parameter $\theta$ with policy gradient methods \cite{sutton2000policy}. SplitNet is trained to maximize the following objective function:
\begin{align}\label{eq:policy_gradient}
\mathcal{L}_{\rm{rl}}(\theta)\!=\mathbb{E}_{(\mathbf{x}, \mathbf{y}, \mathbf{z}){\sim}\mathcal{D}}[\mathbb{E}_{\tilde{\mathbf{a}}{\sim}p_{\theta}(\tilde{\mathbf{a}}|\mathbf{x},\mathbf{y})}R(\tilde{\mathbf{a}},\mathbf{z})].
\end{align}
In practice, REINFORCE algorithm~\cite{williams1992simple} is applied to approximate Equation~\ref{eq:policy_gradient} via sampling $\tilde{\mathbf{a}}$ from $p_{\theta}(\tilde{\mathbf{a}}|\mathbf{x},\mathbf{y})$ for $M$ times, where $M$ is a hyper-parameter representing the sample size. Furthermore, subtracting a baseline \cite{weaver2001optimal} on $R(\tilde{\mathbf{a}},\mathbf{z})$ is also applied to reduce variance. The final objective function is as follows:
\begin{align}
    \mathcal{L}_{\rm{rl}}(\theta)\!\!=\!\!\mathbb{E}_{(\mathbf{x}, \mathbf{y}, \mathbf{z}){\sim}\mathcal{D}}&\Big[\!\sum\limits_{i=1}^{M} p_{\theta}(\tilde{\mathbf{a}}_{i}|\mathbf{x},\mathbf{y})\!\big(R(\tilde{\mathbf{a}}_{i},\mathbf{z})\!\!-\!\!{\bar{R}}\big)\!\Big],\nonumber \\
    \text{where}\ &\bar{R}\!=\! \frac{1}{M}\sum\limits_{i=1}^{M}R(\tilde{\mathbf{a}}_{i},\mathbf{z}).
\end{align}

\subsection{Phase II: Recombine}\label{SRS}

Here we present Phase II with two questions: (1) Receiving the sampled label sequence $\tilde{\mathbf{a}}$, how to compute its reward $R(\tilde{\mathbf{a}},\mathbf{z})$; (2) How to do training and inference for $P_{\rm{rec}}$.

\subsubsection{Reward Computation}

Receiving the label sequence $\tilde{\mathbf{a}}$, we first enumerate all conflicting way candidates. Following the example in Figure \ref{fig:span_restate}, once we get a deterministic $\tilde{\mathbf{a}}$, the split of $(\mathbf{x},\mathbf{y})$ is uniquely determined. Here $\mathbf{x}$ and $\mathbf{y}$ are split into $3$ and $2$ spans respectively. Treating spans as units, we enumerate all conflicting way candidates methodically. We act up to the one-to-one conflicting principle, which means a span either has no conflict (denoted as \emph{EMPTY}) or has only one conflict with a span in another query. Let $\mathcal{C}$ denote the set of all conflicting way candidates, the size of which is $13$ in Figure \ref{fig:span_restate}.

For each conflicting way, we deterministically generate a restated query via the process named \textbf{Restatement}. In general, we simply replace spans in the precedent query with their conflicting spans to generate the restated query. For example, in Figure \ref{fig:span_restate}, the first one in $\mathcal{C}$ is restated as ``How about Bill Collins earned''. For spans in the follow-up query, if they contain column names or cell values and do not have any conflict, they are appended to the tail of the precedent query. It is designed to remedy the sub-query situation where there is no conflict (e.g. ``Which opponent received over 537 attendance'' and ``And which got the result won 5-4''). Specially, if a span in the follow-up query contains a pronoun, we will in reverse replace it with its conflicting span to obtain the restated query.

Finally, the reward can be computed. Here we use BLEU and SymAcc\footnote{Their definitions along with the motivations of using them will be explained in Section \ref{sec:metric}.} to build the reward function, expanding $r(\mathbf{z},\tilde{\mathbf{z}})$ in Equation~\ref{eq:rs} as:
\begin{equation}
    \!r(\mathbf{z},\tilde{\mathbf{z}})=\alpha \cdot \text{BLEU}(\mathbf{z},\tilde{\mathbf{z}}) + \beta \cdot \text{SymAcc}(\mathbf{z},\tilde{\mathbf{z}}),\!
\end{equation}
where $\alpha,\beta>0$ and $\alpha + \beta = 1$. The reward for $\tilde{\mathbf{a}}$ can be obtained using Equation~\ref{eq:rs}.

\subsubsection{Training and Inference}
Besides the reward computation, the recombination model $P_{\rm{rec}}$ needs to be trained to maximize Equation~\ref{eq:rs}. To achieve this, we define a conflicting probability matrix $\mathbf{F}\in\mathbb{R}^{N_x\times N_y}$, where $N_x$ and $N_y$ denote the number of spans in $\mathbf{x}$ and $\mathbf{y}$ respectively. The entry $\mathbf{F}_{u,v}$, the conflicting probability between the $u$-th span in $\mathbf{x}$ and the $v$-th span in $\mathbf{y}$, is obtained by normalizing the cosine similarity between their representations. Here the span representation is the subtraction representation \cite{wang2016graph, cross2016span}, which means that span $(x_i,\dots,x_k)$ is represented by $[\overrightarrow{\mathbf{h}_{k}}{-}\overrightarrow{\mathbf{h}_{i}};\overleftarrow{\mathbf{h}_{i}}{-}\overleftarrow{\mathbf{h}_{k}}]$ from the same BiLSTM in the context layer in Section \ref{sec:state}. Given a conflicting way denoted as $\Tilde{c}\in \mathcal{C}$, the probability of generating its corresponding $\Tilde{\mathbf{z}}$ can be written as the multiplication over $g(u,v)$:
\begin{equation}
\vspace{-1.5mm}
P_{\rm{rec}}(\tilde{\mathbf{z}}|\tilde{\mathbf{a}}) = P(\Tilde{c}|\mathbf{F}) = \prod\limits_{u=1}^{N_x} \prod\limits_{v=1}^{N_y} g(u,v),
\end{equation}
where $g(u,v)=\mathbf{F}_{u,v}$ if the $u$-th span in $\mathbf{x}$ conflicts with the $v$-th span in $\mathbf{y}$; otherwise, $g(u,v)=1-\mathbf{F}_{u,v}$. With the above formulation, we can maximize Equation~\ref{eq:rs} through automatic differentiation. To reduce the computation, we only maximize $P_{\rm{rec}}(\tilde{\mathbf{z}}^*|\tilde{\mathbf{a}})$, the near-optimal solution to Equation~\ref{eq:rs}, where $\tilde{\mathbf{z}}^*\!=\!\mathop{\arg\max}\nolimits_{\tilde{\mathbf{z}}\in\mathcal{Z}}(r(\mathbf{z},\tilde{\mathbf{z}}))$ denotes the best predicted restated query so far.

Guided by the golden restated query $\mathbf{z}$, in training, we find out $\tilde{\mathbf{z}}^*$ by computing the reward of each candidate. However in inference, where there is no golden restate query, we can only obtain $\tilde{\mathbf{z}}^*$ from $\mathbf{F}$. Specially, for the $v$-th span in the follow-up query, we find $u^*=\mathop{\arg\max}\nolimits_{\,u} \mathbf{F}_{u,v}$. That means, compared to other spans in the precedent query, the $u^*$-th span has the highest probability to conflict with the $v$-th span in the follow-up query. Moreover, similar to \citet{Lee_2017}, if $\mathbf{F}_{u^*,v}<\lambda$, then the $v$-th span in the follow-up query has no conflict. The hyper-parameter $\lambda>0$ denotes the threshold.

\begin{table*}[t]
    \centering
        \scalebox{0.85}{
		\begin{tabular}{lllllc}
			\toprule 
			\multicolumn{1}{c}{\textbf{Model}} &
			\multicolumn{2}{c}{\textbf{Dev}} & \multicolumn{3}{c}{\textbf{Test}} \\
			\cmidrule(lr){2-3}
            \cmidrule(lr){4-6}
			&
			\multicolumn{1}{c}{\small SymAcc (\%)} & \multicolumn{1}{c}{\small BLEU (\%)} &
			\multicolumn{1}{c}{\small SymAcc (\%)} & \multicolumn{1}{c}{\small BLEU (\%)} & 
			\multicolumn{1}{c}{\small AnsAcc (\%)} \\
			\midrule
            $\textsc{Seq2Seq}^{\dagger}$~\cite{Bahdanau2014NeuralMT} &~~0.63 {\small$\pm$ 0.00} &21.34 {\small$\pm$ 1.14 } &~~0.50 {\small$\pm$ 0.22} &20.72 {\small$\pm$ 1.31 }&-- \\
            $\textsc{CopyNet}^{\dagger}$ \cite{Gu_2016} &17.50 {\small$\pm$ 0.87}&43.36 {\small$\pm$ 0.54} & 19.30 {\small$\pm$ 0.93} &43.34 {\small$\pm$ 0.45} &-- \\ 
            \textsc{Copy+BERT} \cite{devlin2018bert} & 18.63 {\small$\pm$ 0.61}& 45.14 {\small$\pm$ 0.68} &22.00 {\small$\pm$ 0.45}&  44.87 {\small$\pm$ 0.52}&-- \\ 
            $\textsc{Concat}^{\dagger}$ & ~~~~~~~~-- & ~~~~~~~~-- &22.00 {\small$\pm$ ~~--} ~~~~  &52.02 {\small$\pm$   ~~--} ~~~~ &25.24 \\
			$\textsc{E2ECR}^{\dagger}$ \cite{Lee_2017} & ~~~~~~~~-- & ~~~~~~~~-- &27.00 {\small$\pm$ ~~--} ~~~~  &52.47 {\small$\pm$ ~~--} ~~~~ &27.18\\
			$\textsc{FAnDa}^{\dagger}$ \cite{liu2019fanda} & 49.00 {\small$\pm$ 1.28} &60.14 {\small$\pm$ 0.98} &47.80 {\small$\pm$ 1.14}  &59.02 {\small$\pm$ 0.54} &60.19 \\
			\midrule
			\textsc{StAR}  &\textbf{55.38} {\small$\pm$ 1.21}  &\textbf{67.62} {\small$\pm$ 0.65} & \textbf{54.00} {\small$\pm$ 1.09}  & \textbf{67.05} {\small$\pm$ 1.05} & \textbf{65.05} \\ 
            \bottomrule
		\end{tabular}
		}
	\caption{SymAcc, BLEU and AnsAcc on the FollowUp dataset. Results marked ${\dagger}$ are from~\citet{liu2019fanda}.}\label{tab:results_in_followup}
\end{table*}

\subsection{Extension}\label{subsec:ext}
So far, we have introduced the whole process of \textsc{StAR}. Next we explore its extensibility. As observed, when the annotations are restated queries, \textsc{StAR} is parser-independent and can be incorporated into any context-independent semantic parser. But what if the annotations are answers to follow-up queries? Assuming we have an ideal semantic parser, a predicted restated query $\Tilde{\mathbf{z}}$ can be converted into its corresponding answer $\Tilde{w}$. For example, given $\Tilde{\mathbf{z}}$ as ``where are the players from'', $\Tilde{w}$ could be ``Las Vegas''. Therefore, revisiting Equation \ref{eq:lrl}, in theory \textsc{StAR} is able to be extended by redesigning $r$ as $r(w,\Tilde{w})$, where $w$ denotes the answer annotation. We conduct an extension experiment to verify it, as discussed in Section \ref{sec:sqa}.

\section{Experiments}\label{sec:exp}

In this section, we demonstrate the effectiveness of \textsc{StAR} on the FollowUp dataset\footnote{\url{http://github.com/SivilTaram/FollowUp}} with restated query annotations, and its promising extensibility on the SQA dataset\footnote{\url{http://aka.ms/sqa}} with answer annotations.

\subsection{Implementation details}\label{sec:implementation_details}
We utilize PyTorch \cite{paszke2017automatic} and AllenNLP \cite{gardner2018allennlp} for implementation, and adopt Adam \cite{kingma2014adam} as the optimizer. The dimensions of word embedding and hidden state are both $100$. Variational dropout \cite{kingma2015variational} is employed at embedding layer for better generalization ability (with probability $0.5$). The learning rate is set to be $0.001$ for pre-training, $0.0001$ for RL training on FollowUp, and $0.0002$ for SQA. In the implementation of the REINFORCE algorithm, we set $M$ to be $20$. Finally, for hyper-parameters, we set $\alpha=0.5$, $\beta=0.5$ and $\lambda=0.6$. All the results are averaged over $5$ runs with random initialization.

\subsection{Results on FollowUp dataset}\label{sec:metric}

The FollowUp dataset contains $1000$ natural language query triples $(\mathbf{x},\mathbf{y},\mathbf{z})$. Each triple belongs to a single database table, and there are $120$ tables in several different domains. Following the previous work, we split them into the sets of size $640$/$160$/$200$ for train/dev/test. We evaluate the methods using both answer level and query level metrics. \emph{AnsAcc} is to check the answer accuracy of predicted queries manually. Concretely, $103$ golden restated queries can be successfully parsed by \textsc{Coarse2Fine} \cite{dong2018coarse}. We parse their corresponding predicted queries into SQL using \textsc{Coarse2Fine} and manually check the answers. Although AnsAcc is most convincing, it cannot cover the entire test set. Therefore, we apply two query level metrics: \emph{SymAcc} detects whether all the SQL-related words are correctly involved in the predicted queries, for example column names, cell values and so on. It reflects the approximate upper bound of AnsAcc, as the correctness of SQL-related words is a prerequisite of correct execution in most cases; \emph{BLEU}, referring to the cumulative 4-gram BLEU score, evaluates how similar the predicted queries are to the golden ones \cite{papineni2002bleu}. SymAcc focuses on limited keywords, so we introduce BLEU to evaluate quality of the entire predicted query.

\subsubsection{Model Comparison}
Our baselines fall into two categories. \textbf{Generation}\textbf{-based}\textbf{ methods} conform to the architecture of sequence-to-sequence ~\cite{sutskever2014sequence} and generate restated queries by decoding each word from scratch. \textsc{Seq2Seq} \cite{Bahdanau2014NeuralMT} is the sequence-to-sequence model with attention, and \textsc{CopyNet} further incorporates a copy mechanism. \textsc{Copy+BERT} incorporates the latest pre-trained BERT model \cite{devlin2018bert} as the encoder of \textsc{CopyNet}. \textbf{Rewriting-based methods} obtain restated queries by rewriting precedent and follow-up queries. \textsc{Concat} directly concatenates the two queries. \textsc{E2ECR} \cite{Lee_2017} obtain restated queries by performing coreference resolution in follow-up queries. \textsc{FanDa} \cite{liu2019fanda} utilizes a structure-aware model to merge the two queries. Our method \textsc{StAR} also belongs to this category.

\paragraph{Answer Level} Table~\ref{tab:results_in_followup} shows AnsAcc results of competitive baselines on the test set. Compared with them, \textsc{StAR} achieves the highest, $65.05\%$, which demonstrates its superiority. Meanwhile, it verifies the feasibility of follow-up query analysis in cooperating with context-independent semantic parsing. Compared with~\textsc{Concat}, our approach boosts over $39.81\%$ on~\textsc{Coarse2Fine} for the capability of context-dependent semantic parsing.

\paragraph{Query Level} Table \ref{tab:results_in_followup} also shows SymAcc and BLEU of different methods on the dev and test sets. As observed, \textsc{StAR} significantly outperforms all baselines, demonstrating its effectiveness. For example, \textsc{StAR} achieves an absolute improvement of $8.03\%$ BLEU over the state-of-the-art baseline \textsc{FanDa} on testing. Moreover, the rewriting-based baselines, even the simplest \textsc{Concat}, perform better than the generation-based ones. It suggests that the idea of rewriting is more reasonable for the task, where precedent and follow-up queries are of full utilization.

\begin{table}[t]
    \centering
     \scalebox{0.89}{
		\begin{tabular}{lcc}
			\toprule 
			\textbf{Variant}  & {\small\textbf{SymAcc (\%)}} & {\small \textbf{BLEU (\%)}}\\
			\midrule
			\textsc{StAR}  & 55.38 & 67.62\\
			\midrule
			-- Phase I  & 40.63 & 61.82\\
			-- Phase II  & 23.12 & 48.65\\
			-- RL & 41.25 & 60.19 \\
			+ Basic Reward  & 43.13 & 58.48\\
			+ Oracle Reward  & 45.20 & 63.04\\ 
			+ Uniform Reward  & 53.40 & 66.93\\
            \bottomrule
		\end{tabular}
	}
	\caption{Variant results on FollowUp dev set.}\label{tab:ablation_results}
\end{table}

\subsubsection{Variant Analysis}\label{subsec:variant}
\begin{figure}
    \centering
    \includegraphics[width=.45\textwidth]{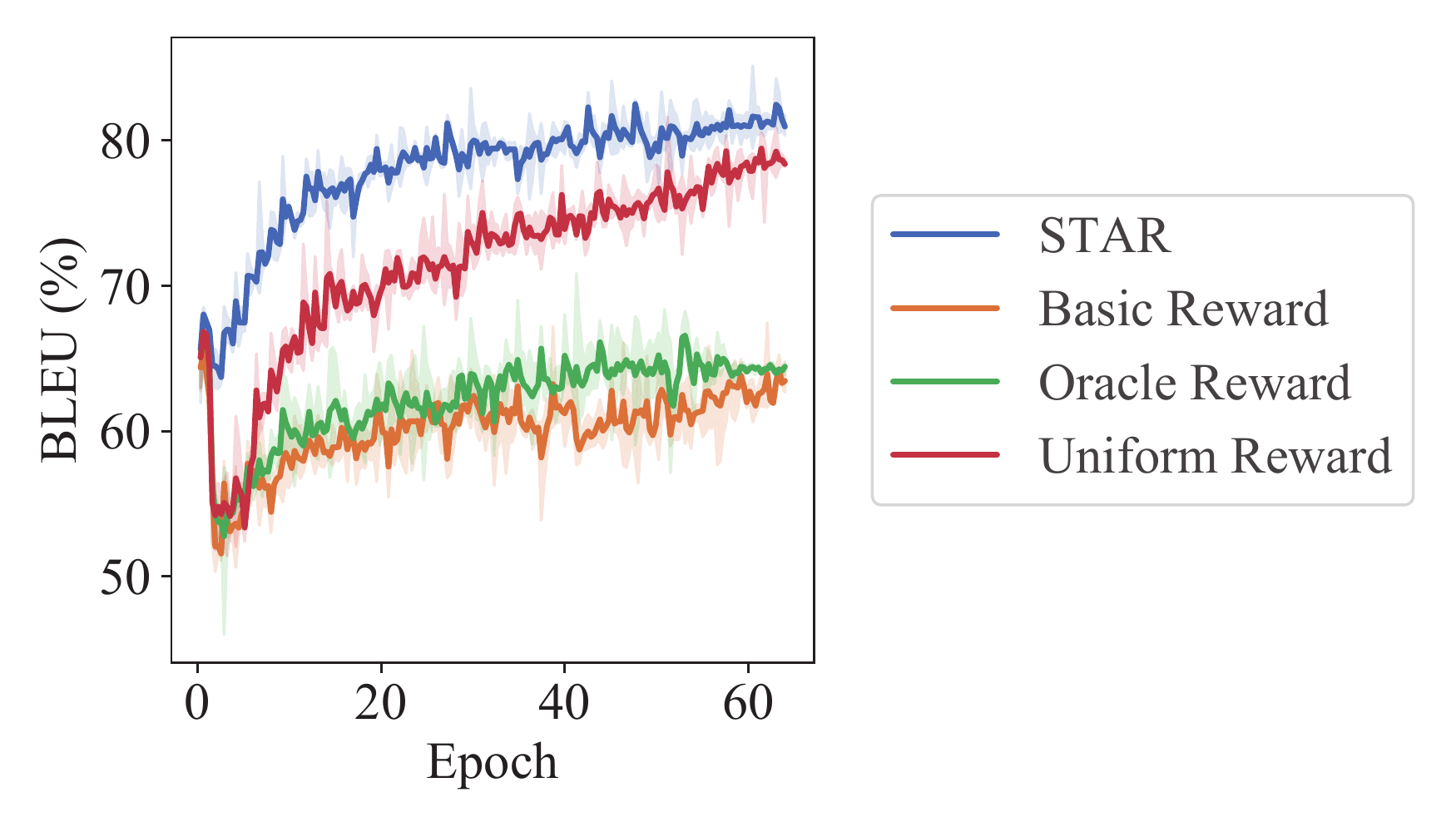}
    \caption{Learning curve on FollowUp train set.}
    \label{fig:rl_learning}
\end{figure}

Besides baselines, we also conduct experiments with several variants of \textsc{StAR} to further validate the design of our model. As shown in Table \ref{tab:ablation_results}, there are three variants with ablation: ``-- Phase I'' takes out SplitNet and performs Phase II on word level; ``-- Phase II'' performs random guess in the recombination process for testing; and ``-- RL'' only contains pre-training. The SymAcc drops from about $55\%$ to $40\%$ by ablating Phase I, and to $23\%$ by ablating Phase II. Their poor performances indicate both of the two phases are indispensable. ``-- RL'' also performs worse, which again demonstrates the rationality of applying RL.

\begin{table*}[t]
    \centering
    \scalebox{1}{
        \begin{tabular}{c}
            \includegraphics[width=.9\textwidth]{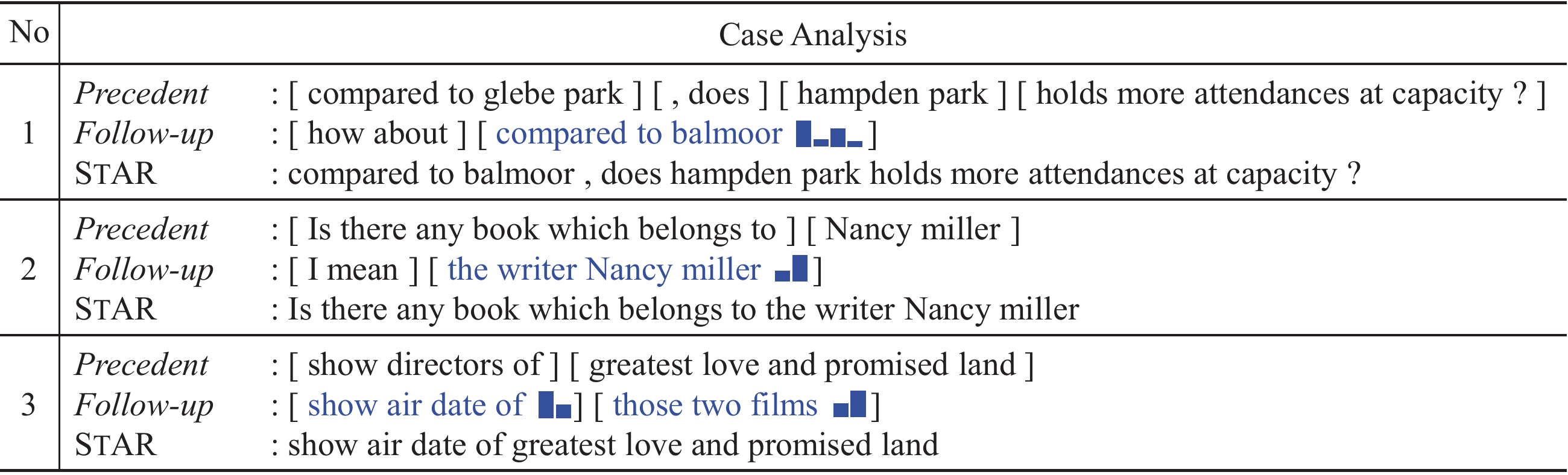}
        \end{tabular}
    }
	\caption{Case analysis of \textsc{StAR} on FollowUp dataset. Square brackets denote different spans.}\label{table:case_study}
\end{table*}

Three more variants are presented with different designs of $R(q,\mathbf{z})$ to prove the efficiency and effectiveness of Equation \ref{eq:rs} as a reward. ``+ Basic Reward'' represents the most straightforward REINFORCE algorithm, which samples both $q\in\mathcal{Q}$ and $\Tilde{\mathbf{z}}{\in}\mathcal{Z}$, then takes $r(\mathbf{z},\Tilde{\mathbf{z}})$ as $R(q,\mathbf{z})$. ``+ Oracle Reward'' assumes the conflicts are always correct and rewrites $R(q,\mathbf{z})$ as $\max_{\tilde{\mathbf{z}}{\in}\mathcal{Z}}(r(\mathbf{z},\tilde{\mathbf{z}}))$. ``+ Uniform Reward'' assigns the same probability to all $\Tilde{\mathbf{z}}$ and obtains $R(q,\mathbf{z})$ as ${\rm mean}({{\sum_{\tilde{\mathbf{z}}{\in}\mathcal{Z}}}}r(\mathbf{z},\tilde{\mathbf{z}}))$. As shown in Table \ref{tab:ablation_results} and Figure \ref{fig:rl_learning}, \textsc{StAR} learns better and faster than the variants due to the reasonable reward design. In fact, as mentioned in Section \ref{sec-overview}, the vast action space of the most straightforward REINFORCE algorithm leads to poor learning. \textsc{StAR} shrinks the space from $|\mathcal{Q}|\cdot|\mathcal{Z}|$ down to $|\mathcal{Q}|$ by enumerating $\tilde{\mathbf{z}}$. Meanwhile, statistics show that \textsc{StAR} obtains a $15\times$ speedup over ``+ Basic Reward'' on the convergence time.

\begin{figure}
    \centering
    \includegraphics[width=.43\textwidth]{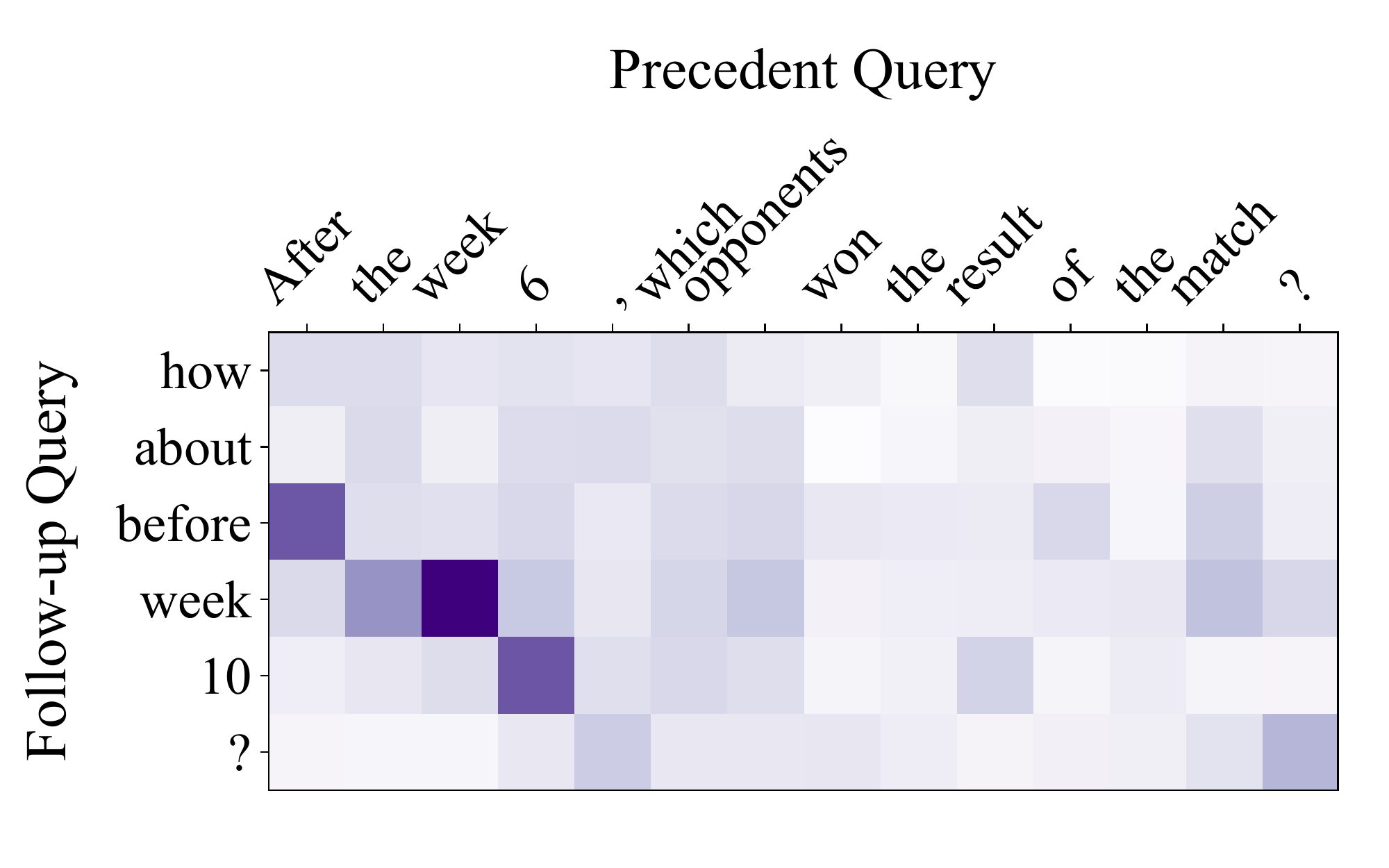}
    \caption{An example of similarity matrix in SplitNet.}
    \label{fig:attn_matrix}
\end{figure}

\subsubsection{Case Study}

Figure \ref{fig:attn_matrix} shows a concrete example of the similarity matrix $\mathbf{A}$ on attention layer of SplitNet. The span ``before week 10'' is evidently more similar to ``After the week 6'' than to others, which meets our expectations. Moreover, the results of three real cases are shown in Table \ref{table:case_study}. The spans in blue are those have conflicts, and the histograms represent the conflict probabilities to all the spans in precedent queries. In Case 1, ``glebe park'', ``hampden park'' and ``balmoor'' are all cell values in the database table with similar meanings. \textsc{StAR} correctly finds out the conflict between ``compared to glebe park'' and ``compared to balmoor'' with the highest probability. Case 2 shows \textsc{StAR} can discover the interrelation of words, where ``the writer Nancy miller'' is learnt as a whole span to replace ``Nancy miller'' in the precedent query. As for Case 3, \textsc{StAR} successfully performs coreference resolution and interprets ``those two films'' as ``greatest love and promised land''. Benefiting from two phases, \textsc{StAR} is able to deal with diverse follow-up scenarios in different domains.

\subsubsection{Error Analysis}

Our approach works well in most cases except for few ones, where SplitNet fails. For example, given the precedent query ``what's the biggest zone?'' and the follow-up query ``the smallest one'', \textsc{StAR} prefers to recognize ``the biggest zone'' and ``the smallest one'' as two spans, rather than perform split operations inside them. The SplitNet fails probably because the conflicting spans, ``the biggest'' $\leftrightarrow$ ``the smallest'' and ``zone'' $\leftrightarrow$ ``one'', are adjacent, which makes it difficult to identify span boundaries well.

\subsection{Extension on SQA dataset}\label{sec:sqa}

\begin{table}[t]
    \centering
    \scalebox{0.9}{
		\begin{tabular}{lcc}
			\toprule 
			\multicolumn{1}{c}{\textbf{Model}} & \!\!	\!\!{\small Precedent} &\!\! {\small Follow-up} \\
			\midrule
            \!\!DynSP~\cite{iyyer2017search} &  70.9  &  35.8 \\
            \!\!NP~\cite{neelakantan2015neural} &  58.9 & 35.9 \\
            \!\!NP + \textsc{StAR} & 58.9 & 38.1 \\
            \!\!DynSP + \textsc{StAR} & 70.9 & 39.5 \\
            \!\!DynSP$^*$~\cite{iyyer2017search} & 70.4 & 41.1 \\
            \bottomrule
		\end{tabular}
		}
	\caption{Answer accuracy on SQA test set.}\label{tab:sqa_results}
\end{table}

Finally, we demonstrate \textsc{StAR}'s extensibility in working with different annotations. As mentioned in Section~\ref{subsec:ext}, by designing $r(w, \Tilde{w})$, \textsc{StAR} can cooperate with the answer annotations. We conduct experiments on the SQA dataset, which consists of $6066$ query sequences ($5042$/$1024$ for train/test). Each sequence contains multiple natural language queries and their answers, where we are only interested in the first query and the immediate follow-up one. As discussed in \cite{iyyer2017search}, every answer can be represented as a set of cells in the tables, each of which is a multi-word value, and the intentions of the follow-up queries mainly fall into three categories. \emph{Column selection} means the follow-up answer is an entire column; \emph{Subset selection} means the follow-up answer is a subset of the precedent answer; and \emph{Row selection} means the follow-up answer has the same rows with the precedent answer.

We employ two context-independent parsers, DynSP \cite{iyyer2017search} and NP \cite{neelakantan2015neural}, which are trained on the SQA dataset to provide relatively reliable answers for reward computing. Unfortunately, they both perform poorly for the restated queries, as the restated queries are quite different from the original queries in SQA. To address the problem, we redesign the recombination process. Instead of generating the restated query, we recombine the predicted precedent answer $\Tilde{w}_{x}$ and the predicted follow-up answer $\Tilde{w}_{y}$ to produce the restated answer $\Tilde{w}$. Therefore, the objective of Phase II is to assign an appropriate intention to each follow-up span via an additional classifier. The goal of Phase I turns to split out spans having obvious intentions such as ``of those''. The way of recombining answer is determined by the voting from intentions on all spans. If the intention column selection wins, then ${\Tilde{w}}={\Tilde{w}_{y}}$; for subset selection, we obtain the subset $\Tilde{w}$ by taking the rows of $\Tilde{w}_{y}$ as the constraint and applying it to $\Tilde{w}_{x}$; and for row selection, we take the rows of $\Tilde{w}_{x}$ and the columns of $\Tilde{w}_{y}$ as the constraints, then apply them to the whole database table to obtain the answer $\Tilde{w}$ retrieved by the predicted SQL. The reward $r(w, \Tilde{w})$ is computed based on Jaccard similarity between the gold answer $w$ and $\Tilde{w}$ as in \cite{iyyer2017search}, and the overall training process remains unchanged.

Table \ref{tab:sqa_results} shows the answer accuracy of precedent and follow-up queries on test set. DynSP$^*$ \cite{iyyer2017search} is designed for SQA by introducing a special action \emph{Subsequent} to handle follow-up queries based on DynSP. DynSP$^*$ is incapable of being extended to work with the annotation of the restated queries. We attempt to apply DynSP$^*$ (trained on SQA) directly on FollowUp test set, which results in an extremely low AnsAcc. On the contrary, \textsc{StAR} is extensible. ``+\textsc{StAR}'' means our method \textsc{StAR} is incorporated into the context-independent parser and empowers them with the ability to perform follow-up query analysis. As observed, integrating \textsc{StAR} consistently improves performance for follow-up queries, which demonstrates the effectiveness of \textsc{StAR} in collaborating with different semantic parsers. The comparable results of DynSP+\textsc{StAR} to DynSP$^*$ further verifies the promising extensibility of \textsc{StAR}.

\section{Related Work}

Our work is closely related to two lines of work: context-dependent sentence analysis and reinforcement learning. From the perspective of context-dependent sentence analysis, our work is related to researches like reading comprehension in dialogue \cite{reddy2018coqa,choi2018quac}, dialogue state tracking \cite{williams2013dialog}, conversational question answering in knowledge base \cite{saha2018complex,guo2018dialog}, context-dependent logic forms \cite{long2016simpler}, and non-sentential utterance resolution in open-domain question answering \cite{raghu2015statistical,kumar2017incomplete}. The main difference is that we focus on the context-dependent queries in NLIDB which contain complex scenarios. As for the most related context-dependent semantic parsing, \citet{zettlemoyer2009learning} proposes a context-independent CCG parser and then conduct context-dependent substitution, \citet{iyyer2017search} presents a search-based method for sequential questions, and \citet{suhr2018learning} presents a sequence-to-sequence model to solve the problem. Compared to their methods, our work achieves context-dependent semantic parsing via learnable restated queries and existing context-independent semantic parsers.

Moreover, the technique of reinforcement learning has also been successfully applied to natural language tasks in dialogue, such as hyper-parameters tuning for coreference resolution \cite{clark2016deep}, sequential question answering \cite{iyyer2017search} and coherent dialogue responses generation \cite{li2016deep}. In this paper, we employ reinforcement learning to capture the structures of queries, which is similar to \citet{zhang2018learning} for text classification.

\section{Conclusion and Future Work}
We present a novel method, named Split-And-Recombine (\textsc{StAR}), to perform follow-up query analysis. A two-phase process has been designed: one for splitting precedent and follow-up queries into spans, and the other for recombining them. Experiments on two different datasets demonstrate the effectiveness and extensibility of our method. For future work, we may extend our method to other natural language tasks.

\section*{Acknowledgments}

We thank all the anonymous reviewers for their valuable comments. This work was supported by the National Natural Science Foundation of China (Grant Nos. U1736217 and 61932003).

\bibliography{emnlp-ijcnlp-2019}
\bibliographystyle{acl_natbib}

\end{document}